\documentclass{article}

    \usepackage[final]{neurips_2024}

\usepackage[utf8]{inputenc} 
\usepackage[T1]{fontenc}    
\usepackage{hyperref}       
\usepackage{url}            
\usepackage{booktabs}       
\usepackage{amsfonts}       
\usepackage{nicefrac}       
\usepackage{microtype}      
\usepackage{xcolor}         
\usepackage{graphicx}
\usepackage{caption}
\usepackage{subcaption}
\usepackage{wrapfig}
\usepackage{tikz}
\usetikzlibrary{shapes.geometric, positioning, arrows.meta, fit}

\usepackage{amssymb}        
\usepackage{amsmath}        
\newtheorem{theorem}{Theorem}
\newtheorem{proof}{Proof}
\setcitestyle{numbers,square}

\title{From Text to Pose to Image: Improving Diffusion Model Control and Quality}

\author{%
    Clément Bonnet\thanks{Correspondence to: <\texttt{clement.bonnet16@gmail.com}>. Work done while at Raive.} \And
    Ariel N. Lee \\ Raive \And
    Franck Wertel \\ Raive \And
    Antoine Tamano \\ Raive \And
    Tanguy Cizain \\ Raive \And
    Pablo Ducru \\ Raive
}

\begin{document}

\maketitle

\begin{abstract}
In the last two years, text-to-image diffusion models have become extremely popular.
As their quality and usage increase, a major concern has been the need for better output control.
In addition to prompt engineering, one effective method to improve the controllability of diffusion models has been to condition them on additional modalities such as image style, depth map, or keypoints.
This forms the basis of ControlNets or Adapters.
When attempting to apply these methods to control human poses in outputs of text-to-image diffusion models, two main challenges have arisen.
The first challenge is generating poses following a wide range of semantic text descriptions, for which previous methods involved searching for a pose within a dataset of (caption, pose) pairs.
The second challenge is conditioning image generation on a specified pose while keeping both high aesthetic and high pose fidelity.
In this article, we fix these two main issues by introducing a text-to-pose (T2P) generative model alongside a new sampling algorithm, and a new pose adapter that incorporates more pose keypoints for higher pose fidelity.
Together, these two new state-of-the-art models enable, for the first time, a generative text-to-pose-to-image framework for higher pose control in diffusion models.
We release all models and the code used for the experiments at \href{https://github.com/clement-bonnet/text-to-pose}{https://github.com/clement-bonnet/text-to-pose}.
\end{abstract}

\section{Introduction}

Text-to-image diffusion models have recently shown impressive results in the space of image generation~\cite{sohl-dickstein_deep_2015, ho2020denoising, song_score-based_2021, karras_elucidating_2022, Rombach_2022_CVPR}.
As the use and quality of these models have increased, so have the requirements from users to strengthen the controllability of the outputs.
Various methods have emerged to yield better output control, from fine-tuning~\cite{xiao_tackling_2022}, to sophisticated prompt-engineering~\cite{gal2022image, meng2022sdedit}, to adding new modalities as conditions, e.g. ControlNets~\cite{zhao_uni-controlnet_2023} and Adapters \cite{mou_t2i-adapter_2023, ye_ip-adapter_2023}.

In the context of enhancing human pose fidelity in text-to-image diffusion models, two key problems have emerged:
\begin{itemize}
    \item (1) Obtaining poses to cover the wide variety of situations that can be described semantically through the text description.
    Previous methods required selecting poses from a dataset, either by giving a picture, extracting its pose using pose-estimation models such as DWPose~\cite{yang2023effective}, and transferring it to the new image with systems such as GANs~\cite{ma_pose_2017}.
    \item (2) Pose-conditioned image generation such as the previous SOTA SDXL-Tencent adaptor~\cite{mou_t2i-adapter_2023} does not include faces or hands, hindering pose fidelity, suffers from an aesthetic quality drastically lower than that of the original SDXL model.
\end{itemize}

In this article, we overcome these two main challenges of image generation conditioned on text and pose by first training a text-to-pose generative model, which is to the best of our knowledge the first of its kind, and then by training a new SOTA pose adapter for diffusion models, which incorporates both facial and hand gestures.
Combining these two new SOTA models enables a new text-to-pose-to-image generative framework for higher pose control in diffusion models (see figure~\ref{fig:intro} for a visual in the appendix).

\section{Text-to-Pose Generative Model}
\label{sec: text-to-pose} 

We describe a human pose with a series of points locating key positions of body parts: 18 points for the body, 42 points for the hands, and 68 points for the face (see Figure \ref{fig:clapp_heatmap} in the appendix for examples of poses).
To train a text-to-pose model, we first design a metric inspired by CLIP~\cite{radford2021learning} to assess the quality of generated poses during training.
We then design a transformer architecture for prompt-conditioned pose generation, annotate a dataset of high-quality images, and train the model on it.

\subsection{CLaPP: A Contrastive Text-Pose Metric}

Before training a generative text-to-pose model, we first create a metric that will guide training regarding matching text to poses.
To do so, we train a contrastive model -- akin to CLIP~\cite{radford2021learning,CLIP_ramesh_hierarchical_2022} -- by embedding both text and pose image into a joint latent space and optimizing the projection such as to minimize the angular distance between poses that corresponds to the same text description while maximizing the distance between those that do not.
Inspired by CLIP, we call our model \emph{Contrastive Language-Pose Pretraining} (CLaPP).

We train our CLaPP metric on a dataset of 500k (image, prompt) pairs from JourneyDB~\cite{sun2023journeydb} which we annotate to extract poses (body, face, and hands) using DWPose~\cite{yang2023effective}.
Both prompts and poses (images of poses on a black background)  are first encoded using CLIP.
The CLaPP layers then map each CLIP representation to a new text-pose embedding from which to compute the CLaPP score.
The resulting CLaPP scores between different poses and texts are reported in Figure \ref{fig:clapp_heatmap} in the appendix.
As expected, the diagonal of the score matrix has high CLaPP scores since it corresponds to the exact captions of each image.

\subsection{T2P: Text-to-Pose Transformer}

\begin{figure}[ht]
    \centering
    \begin{tikzpicture}[node distance=0.5cm and 0.5cm, 
    auto, 
    >=Latex, 
    block/.style={rectangle, draw, align=center, rounded corners, minimum height=3em},
    textnode/.style={inner sep=2pt, align=center},
    line/.style={draw, ->}] 

    \node[textnode] (pose) {Pose \\ Sequence};
    \node[block, right=of pose] (embedding) {Embedding};
    \node[block, right=of embedding, xshift=0.3cm] (sa) {Causal\\Self-Attention};
    \node[block, right=of sa] (ca) {Cross-\\Attention};
    \node[textnode, above=1cm of ca] (clip) {CLIP Text Features};
    \node[block, right=of ca] (ff) {Feed \\Forward};
    \node[block, right=of ff, xshift=0.3cm] (linear) {Linear};
    \node[block, below=of linear, xshift=-1.6cm] (gmm) {GMM \\ Parameters};
    \node[block, below=of linear, xshift=0.5cm] (classification) {Binary \\ Classification};

    \path[line] (pose) -- (embedding);
    \path[line] (embedding) -- (sa);
    \path[line] (sa) -- (ca);
    \path[line] (clip) -- (ca);
    \path[line] (ca) -- (ff);
    \path[line] (ff) -- (linear);
    \path[line] (linear) -- (gmm);
    \path[line] (linear) -- (classification);

    \node[draw, dashed, rectangle, rounded corners, fit=(sa) (ca) (ff), inner sep=10pt] (loop) {};
    \node[right=0.0cm of loop.north east, anchor=north west, yshift=0.1cm] {xN}; 
\end{tikzpicture}
    \caption{Text-to-Pose transformer architecture.}
    \label{fig:t2p-architecture}
\end{figure}
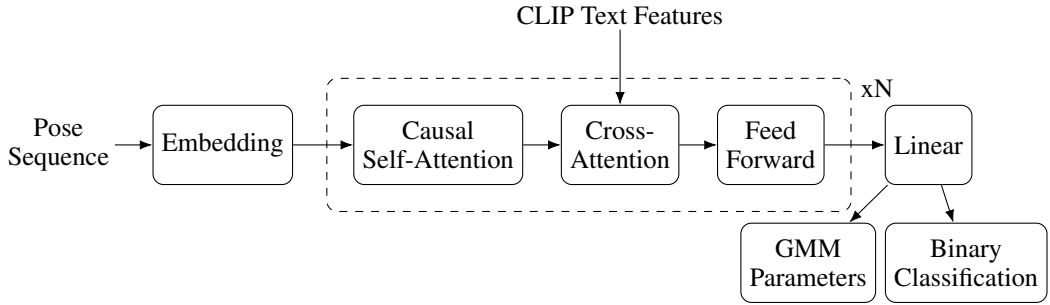

A pose is defined as an ordered sequence of key points, i.e. (x, y) coordinates of points in the image that correspond to e.g. the right thumb, the left shoulder, the nose, etc.
Given this, it makes sense to design a sequence model that can embed the whole pose conditioned on text features.
For this, we use a decoder-only transformer architecture~\cite{vaswani2017attention} (see figure~\ref{fig:t2p-architecture}) to auto-regressively predict the next point in a pose with cross-attention on the text features produced by CLIP.
We use positional embeddings of length 128 (18 body points, 68 face points, and 2$*$21 for the hands).
Contrary to language models that have tokens, the pose input space is continuous, therefore, we model it with a Gaussian Mixture Model (GMM) along with a binary classifier.
The former allows us to learn rather complex non-normal probability distributions over the next pose keypoint, while the latter predicts if the next keypoint exists.
Indeed, poses are not always complete, e.g. if a pose describes a portrait, it is likely that only the 68 points corresponding to the face will be present.
We found the use of mixtures necessary since a single Gaussian cannot represent well the multi-modal properties of the conditional probability distribution of the next keypoint, e.g. if the pose represents two people on either side, a GMM can capture this bi-modality of where the next point can be (left or right).

\subsection{Training}

We train the model in a self-supervised fashion by predicting the next pose point in the sequence.
The GMM output is trained with maximum likelihood (the GMM likelihood can be computed analytically), and the binary classifier with binary cross-entropy on the existence of the next point or not.
We train the model on the 4M (pose, prompt) pairs obtained from JourneyDB by annotating each image with DWPose.
We found that a mixture of 6 different Gaussian mixtures and a transformer with 4 layers worked best for our dataset.

\subsection{Inference: Tempered Distribution Sampling}

After training, the auto-regressive Gaussian mixture model suffers from too high of an entropy which makes sampling from the model quickly diverge out-of-distribution at inference time.
Traditionally, language models have their logits divided by a temperature at inference time to decrease the model's entropy \cite{hinton_distilling_2015, wang_contextual_2020, wang_cost-effective_2023}.
In our case, the model outputs GMM parameters and although the tempered likelihood function can be analytically computed, to our knowledge there is no known algorithm to sample from it.
Therefore, we generalize the approach of tempered distribution sampling to any distribution in theorem~\ref{thm: tempered distribution}.
We then applied tempered sampling to our GMM and found generated poses to be much more precise with lower temperatures, at the expense of being less diverse (see figure~\ref{fig: temperate sampling for text-to-pose} in the appendix).

\subsection{Analysis \& Performance}

\begin{wrapfigure}{r}{0.55\textwidth}
    \centering
    \vspace{-4em}
    \includegraphics[width=0.5\textwidth]{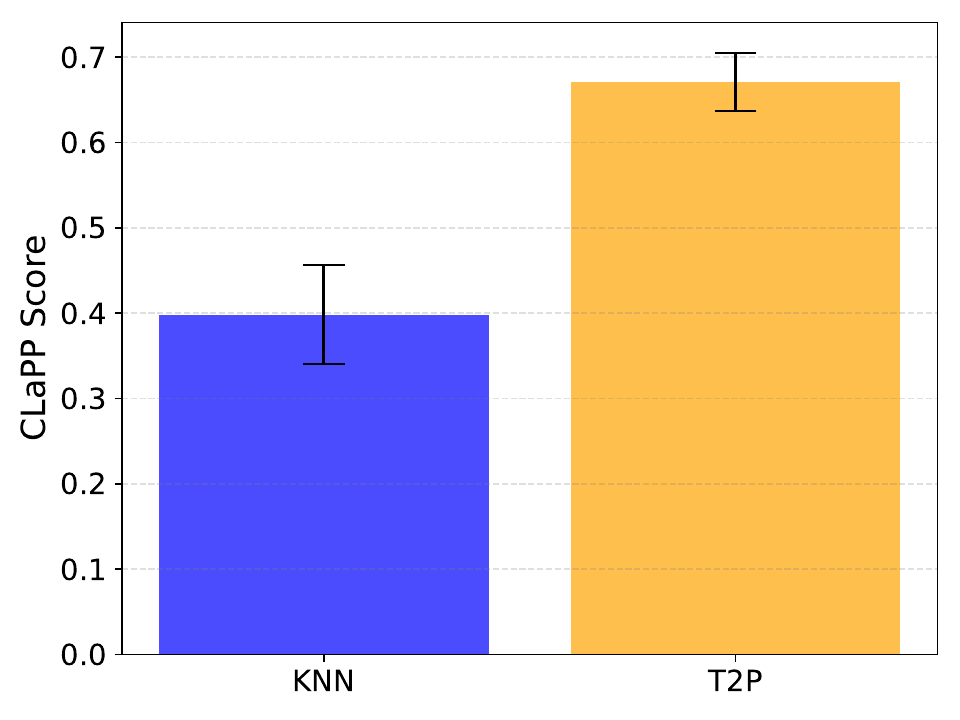}
    \caption{CLaPP scores with 95\% confidence intervals. The win-rate ratio of T2P over KNN is 78\%. We use a subset of 100 (caption, pose) pairs from the COCO 2017 validation dataset.}
    \vspace{-1.5em}
    \label{fig:clapp_scores}
\end{wrapfigure}

To analyze the performance of our text-to-pose (T2P) model, we compiled a ``COCO-Pose'' benchmark dataset, where we extracted the poses of 100 image-caption pairs from the standard COCO 2017 benchmark~\cite{coco}.
For each of the text labels of the COCO-Pose dataset, we then compared the CLaPP score of either: (a) selecting the closest pose from the training dataset (KNN search~\cite{cover_nearest_1967} from CLIP embeddings); or (b) generating a pose using T2P.
The results are compiled in figure~\ref{fig:clapp_scores}, where we can see that T2P outperforms a KNN in the training dataset 78\% of the time.
This somewhat proves the local generalization capabilities of T2P and its alignment with respect to prompts describing poses.

\section{Pose Adapter for Image Generation}
\label{sec: adapter}

Several pose-conditioning systems have been built for diffusion models, notably ControlNets~\cite{zhao_uni-controlnet_2023}, and Adapters~\cite{mou_t2i-adapter_2023, ye_ip-adapter_2023}.
The previous SOTA model for pose-conditioning is the SDXL-Tencent adapter~\cite{mou_t2i-adapter_2023} which suffers from two key drawbacks.
First, it does not include keypoints related to faces and hands. Second, it suffered from lower image aesthetic due to training on low-quality images or optimization issues.

\subsection{Adapter Training}

We start from the same architecture as the previous SOTA adapter~\cite{mou_t2i-adapter_2023}, which consists of a ResNet-like model, and we train it on the same dataset used to train T2P.
We use the same hyper-parameters as the original model for a total of 7600 training steps with a batch size of 256.

\subsection{Analysis \& Performance}

To measure the performance of our adapter compared to the previous SOTA, we conducted a series of benchmarking tests.
For these tests, we generate pose-conditioned images using 30 steps of diffusion with both adapters and then 10 steps of the SDXL refiner model to correct small distortions.
The fact that our adapter now has faces and hands to condition on yields better pose accuracy, as shown in figure \ref{fig: SDXL-Tencent adapter versus ours Dance} in the appendix.
However, one can see the poses are never perfectly matched and the image quality is still below that of the base SDXL model.

\begin{figure}[h!]
    \centering
    \begin{subfigure}[b]{0.32\textwidth}
        \centering
        \includegraphics[width=\textwidth]{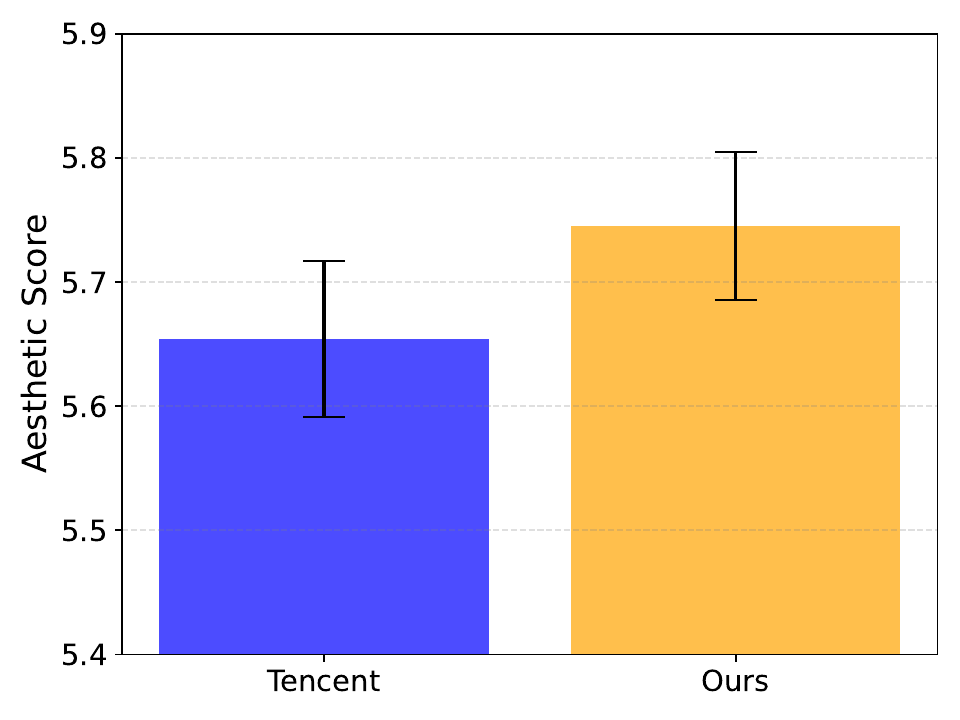}
        \caption{Aesthetic Score, win ratio: 70\%}
        \label{fig:adapter_aesthetic_scores}
    \end{subfigure}
    \hfill
    \begin{subfigure}[b]{0.32\textwidth}
        \centering
        \includegraphics[width=\textwidth]{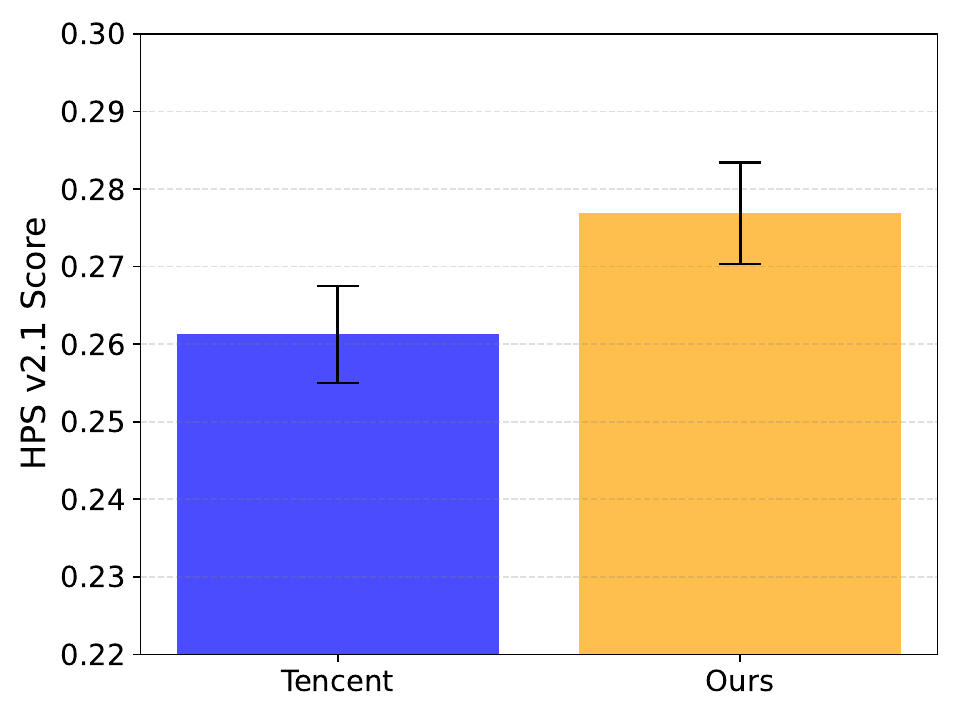}
        \caption{HPS v2, win ratio: 76\%}
        \label{fig:adapter_hps_scores}
    \end{subfigure}
    \hfill
    \begin{subfigure}[b]{0.32\textwidth}
        \centering
        \includegraphics[width=\textwidth]{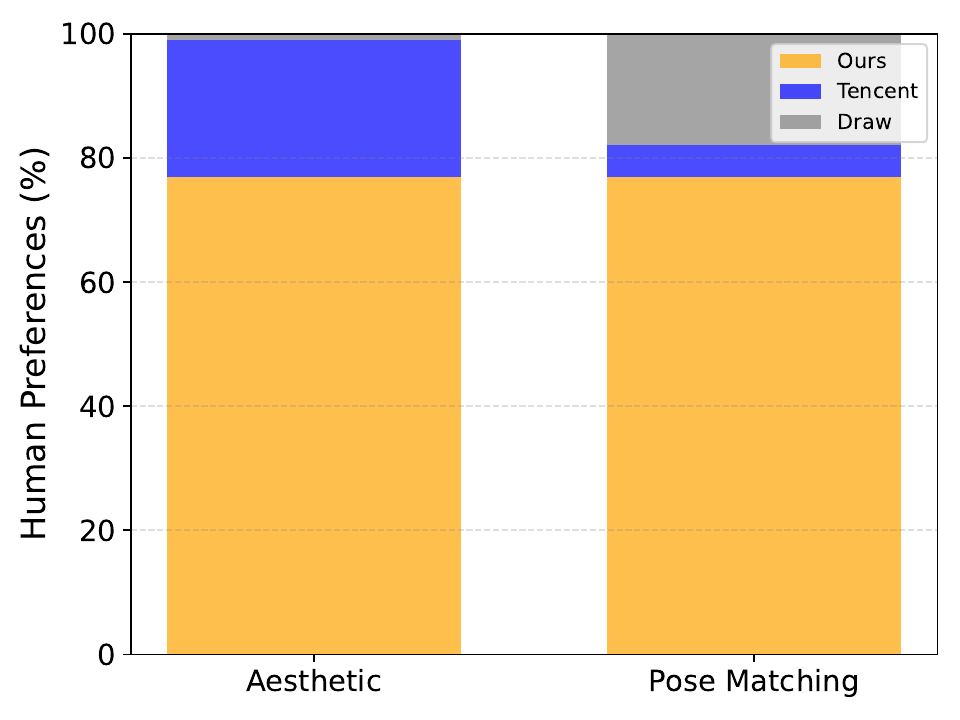}
        \caption{Human Preferences}
        \label{fig:adapter_human_scores}
    \end{subfigure}
    \caption{Performance of pose-conditioned image generation for the Tencent adapter and ours. \textbf{(a)}: Aesthetic score (ML based). \textbf{(b)}: Human Preference Score v2~\cite{wu2023human}. \textbf{(c)}: Human preferences (manually annotating). Error bars represent two standard deviations. We use a subset of 100 (caption, pose) pairs from the COCO 2017 validation dataset to serve as conditions for image generation.}
    \label{fig:adapter_scores_combined}
\end{figure}

We also measure the aesthetic score~\footnote{see \url{https://github.com/christophschuhmann/improved-aesthetic-predictor}}, and Human Preference Score (HPS) v2~\cite{wu2023human}, of both the SDXL-Tencent adapter and ours on the COCO-Pose dataset.
As shown in figure~\ref{fig:adapter_scores_combined}, our adapter outperforms the Tencent one 70\% of the time for the aesthetic score, and 76\% of the time for the HPS score on the COCO-Pose benchmark.

\section{Discussion}

In this article, we tackle the problem of conditioning on human poses the image generation process of diffusion models. 
We first solve the issue of finding the right pose given a prompt by training a text-to-pose auto-regressive model, which outperforms searching the training database (using KNN) 78\% of the time.
We then tackle the task of conditioning on poses in a way that guarantees both high fidelity to the pose and high image aesthetics by training a pose adapter on high-quality images with more pose keypoints than previous models.
Our adapter has a win-rate ratio of 70\% on the aesthetic score and 76\% on the COCO-Pose benchmark compared to the previous SOTA.

This work encourages new paradigms for improved user experiences.
By creating this intermediate modifiable image semantics (human poses), one can imagine slightly altering the pose while keeping the seed constant so as to modify the human pose directly on the generated image, turning the pixel image somewhat vectorized.

Improved image fidelity and human pose control bring our attention to the broader ethical impacts that such technologies can have.
While our system does not guarantee photorealism, it is important to keep in mind that such AI-generated images should not be misused and mixed up with real photos of humans.

\bibliographystyle{plainnat}
\bibliography{references}

\newpage
\appendix
\section{Appendix}

\subsection{Text-to-pose-to-image framework}
\begin{figure}[h]
\centering
\begin{subfigure}{.45\textwidth}
  \centering
  \includegraphics[width=0.9\linewidth]{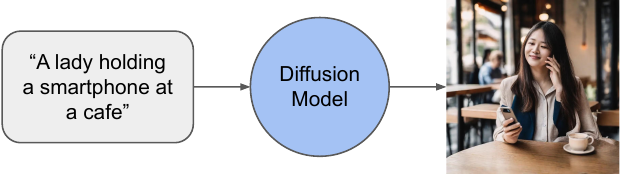}
  \caption{Standard \textit{text-to-image} generation}
  \label{fig:sub1}
\end{subfigure}
\begin{subfigure}{.45\textwidth}
  \centering
  \includegraphics[width=0.9\linewidth]{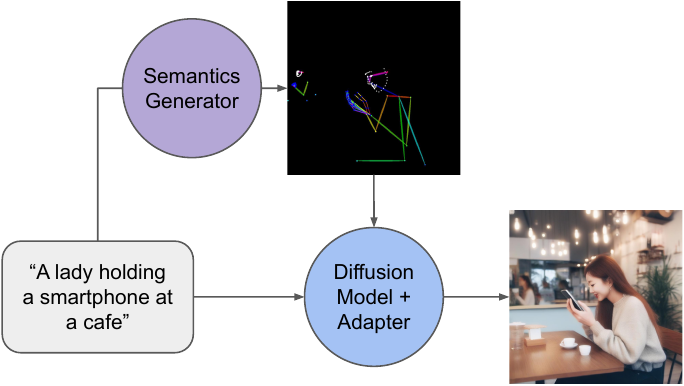}
  \caption{Ours: \textit{text-to-pose-to-image} generation}
  \label{fig:sub2}
\end{subfigure}
\caption{Text-to-pose-to-image framework.}
\label{fig:intro}
\end{figure}

We propose factorizing text-to-image generation into semantics generation (human poses) and then semantics-conditioned image generation (see figure~\ref{fig:intro}).
This allows for better control over the semantics and higher overall quality.

\subsection{CLaPP metric}
\begin{figure}[h]
    \centering
    \includegraphics[width=\textwidth]{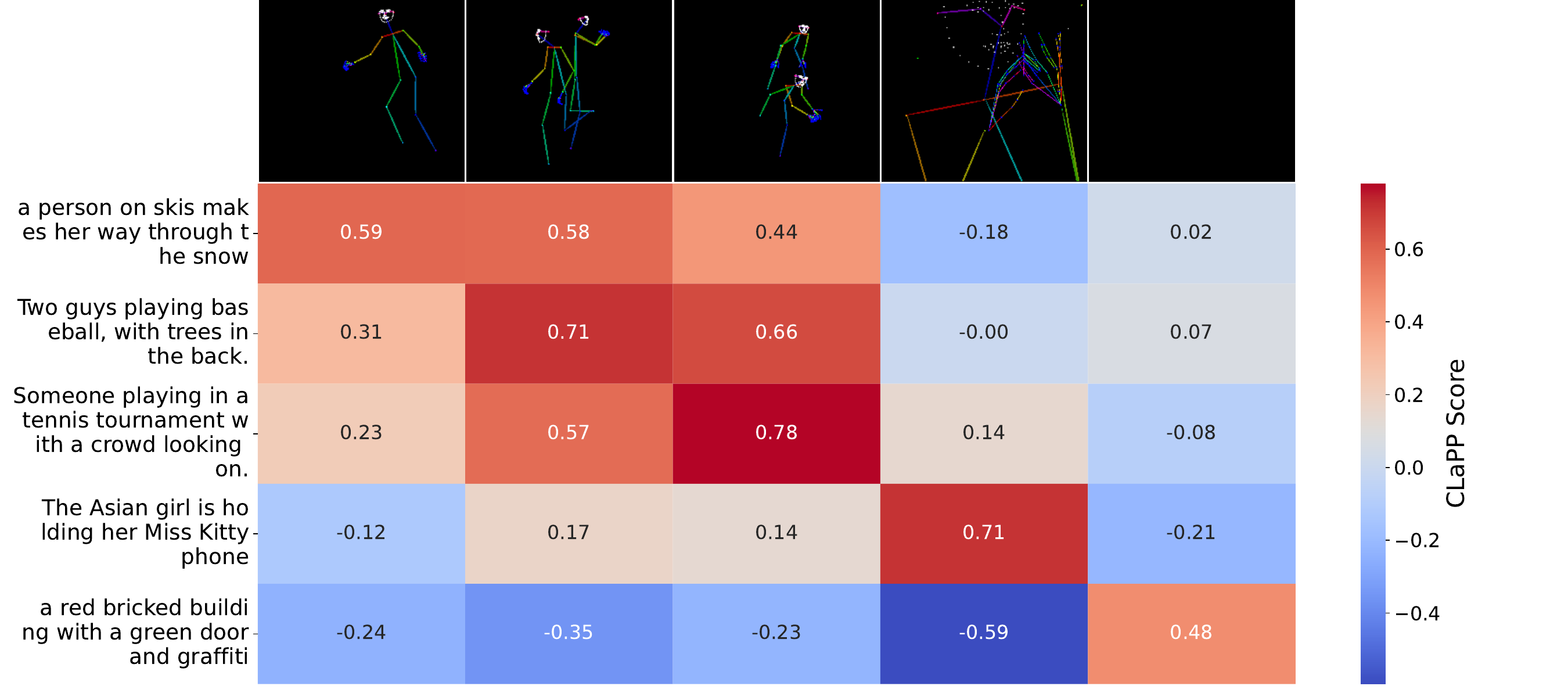}
    \caption{CLaPP scores on 5 poses and corresponding captions from the COCO dataset. The scores measure the compatibility between text and poses.}
    \label{fig:clapp_heatmap}
\end{figure}

Our contrastive model, called CLaPP, can predict compatibility scores between a prompt and a human pose.
Figure~\ref{fig:clapp_heatmap} demonstrates a few CLaPP scores of samples from the COCO dataset.
The closer to 1, the more similar a pose and a prompt are, the worst score being -1.

\subsection{Tempered distribution sampling}
We prove some properties of our generalization of tempered sampling to any distribution in theorem~\ref{thm: tempered distribution}.
We also describe the sampling algorithm used to do inference with the T2P model.

\begin{figure}[h]
    \centering
    \includegraphics[width=0.8\textwidth]{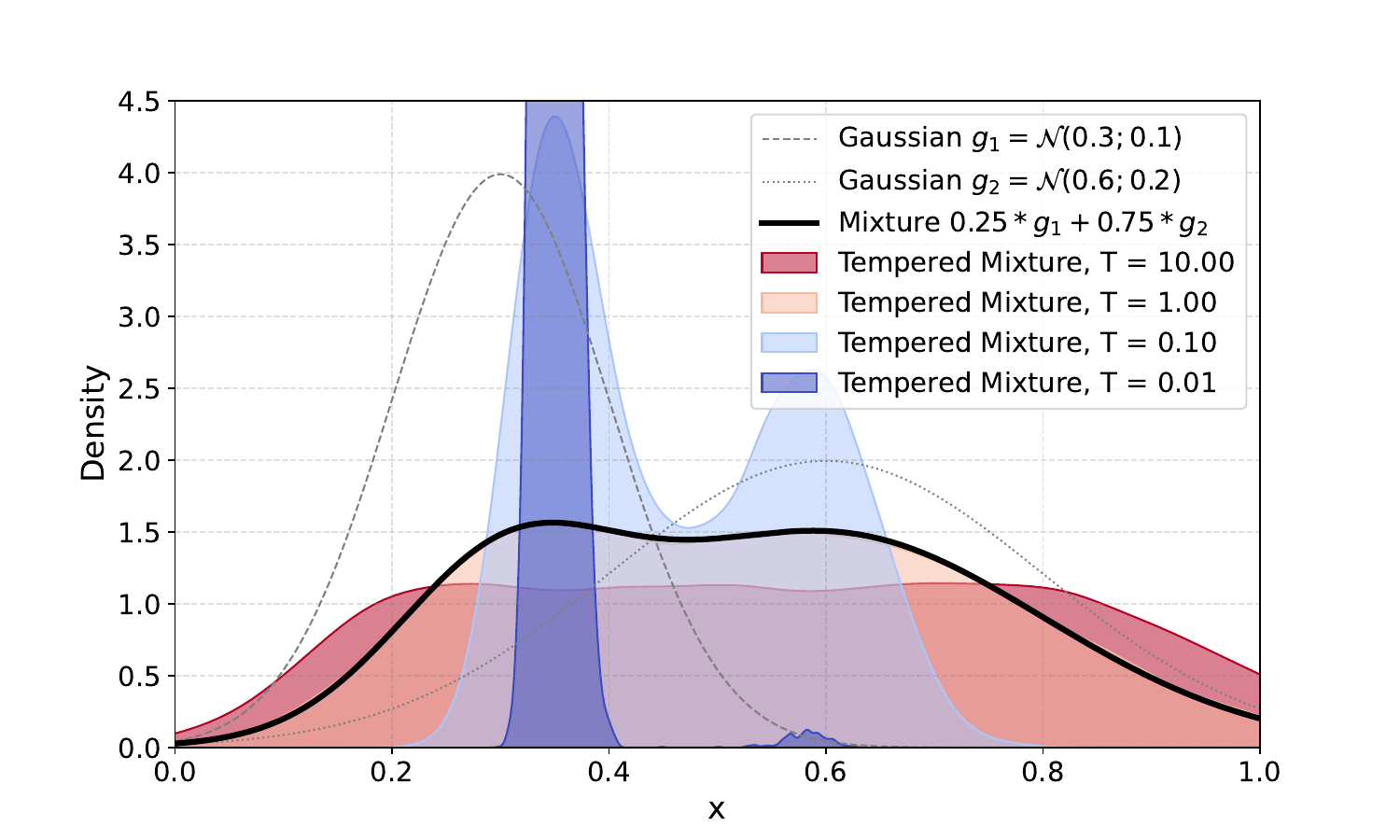}
    \caption{Tempered distribution sampling of Gaussian Mixture Model. $N=10,000$ Monte-Carlo samples.}
    \label{fig:tempered_sampling}
\end{figure}

\begin{theorem}[Tempered distribution sampling]
\label{thm: tempered distribution}
Let $X $ be a random variable, with a probability distribution density $p(X)$, i.e. $ X \sim p(X)$.
Let $T \in \mathbb{R}_+$ be a real positive ``temperature'' parameter.
We define the ``tempered distribution transform'' as $X_T \sim p_T(X)$, where:
\begin{equation}
    p_T(x) \triangleq \frac{p(x)^\frac{1}{T}}{\int_X p(x)^\frac{1}{T} \mathrm{d}x } \ = \ \frac{\mathrm{e}^\frac{\mathrm{ln} \ p(x)}{T}}{\int_X \mathrm{e}^\frac{\mathrm{ln} \ p(x)}{T} \mathrm{d}x } 
    \label{eq: tempered distribution}
\end{equation}

\underline{Properties} -- the tempered distribution $p_T$ is related to the original one $p$ by the following properties:
\begin{itemize}
    \item Conservation of modes (and inflection points): if $\partial_x p = 0$, then $\partial_x p_T = 0$
    \item Temperature one invariance: $p_{T=1}(x) = p(x) $
    \item Mode selection with smaller temperatures: as $T$ goes to zero, the tempered distributions $p_T$ tends towards a dirac distribution selecting the mode of the $p$ distribution, i.e. 
\begin{equation}
p_T(x) \underset{T\to 0}{\longrightarrow} \underset{p(x)}{\mathrm{max}} \; \delta(x)
    \label{eq: mode selecting at low temperatures temperatures}
\end{equation}
    \item Uniform thermalization with high temperatures: as $T$ goes to infinity, the tempered distribution $p_T$ tends towards a uniform distribution over the support of $p$, i.e.
\begin{equation}
p_T(x) \underset{T\to \infty}{\longrightarrow} \underset{X}{\mathrm{Uniform}}(x)
    \label{eq: uniform at high temperatures}
\end{equation}
    \item Temperature score scaling: the score of the tempered distribution $\nabla_x \mathrm{ln} p_T(x)$ is proportional to the score of the original distribution $\nabla_x \mathrm{ln} p(x)$, scaled by the temperature, i.e.
\begin{equation}
\nabla_x \mathrm{ln} p_T(x) = \frac{\nabla_x \mathrm{ln} p(x)}{T}
    \label{eq: score scaling with temperature}
\end{equation}
\end{itemize}

\underline{Sampling scheme} -- to sample from the tempered distribution, one can (c.f. Figure \ref{fig:tempered_sampling}):
\begin{itemize}
    \item Sample $N$ points $x_i$ from $p(X)$ to cover its support
    \item Sample from the following softmax distribution:
\begin{equation}
\mathrm{Softmax} \left\{ \left[ \frac{1}{T} - 1 \right] \mathrm{ln} \ p(x_i) \right\} \ \triangleq \ \frac{ \mathrm{e}^{\left[ \frac{1}{T} - 1 \right] \mathrm{ln} p(x_i)} }{ \sum_{j=1}^N \mathrm{e}^{\left[ \frac{1}{T} - 1 \right] \mathrm{ln} p(x_j)}}
    \label{eq: tempered distribution sampling scheme}
\end{equation}
\end{itemize}

\begin{proof}
The proofs of the mode-conserving and score-scaling properties are established by taking the derivatives from the tempered distribution definitions.
The mode-selecting property is established by taking the temperature to zero.
The proof of the sampling scheme is derived by importance sampling: $\int_X \frac{\mathrm{e}^\frac{\mathrm{ln} \ p(x)}{T}}{p(x)} p(x) \mathrm{d}x = \int_X \mathrm{e}^{\left[ \frac{1}{T} - 1 \right] \mathrm{ln} p(x)} p(x) \mathrm{d}x  $, which is then approximated at large $N$ by Monte Carlo estimate $\sum_{j=1}^N \mathrm{e}^{\left[ \frac{1}{T} - 1 \right] \mathrm{ln} p(x_j)}$. 
\end{proof}
\end{theorem}

We can see in figure~\ref{fig:tempered_sampling} the influence of the temperature on a tempered GMM in a toy case with two Gaussian mixtures.
The closer to 0 the temperature, the closer to the mode the tempered distribution peaks.

\begin{figure}[h]
\centering
\begin{subfigure}{.33\textwidth}
  \centering
  \includegraphics[width=0.9\linewidth]{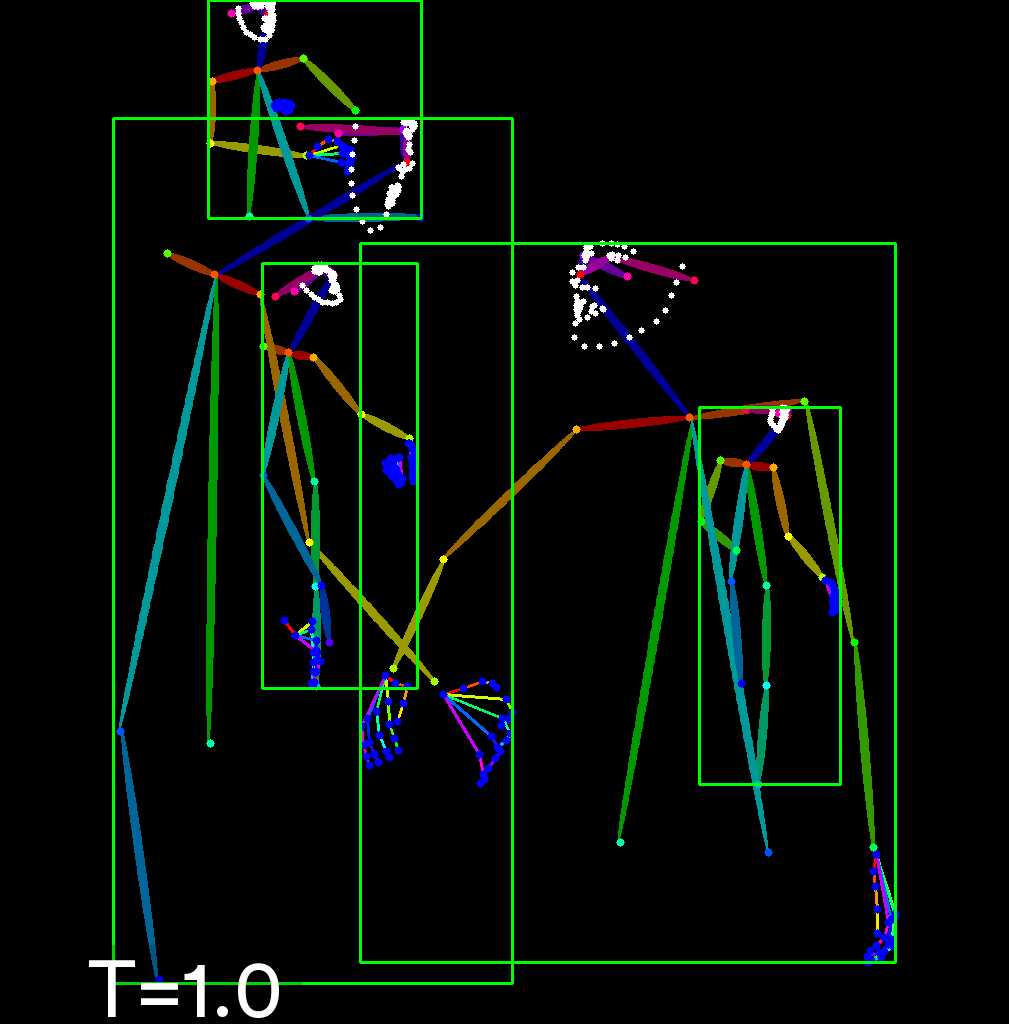}
\end{subfigure}
\begin{subfigure}{.3\textwidth}
  \centering
  \includegraphics[width=0.9\linewidth]{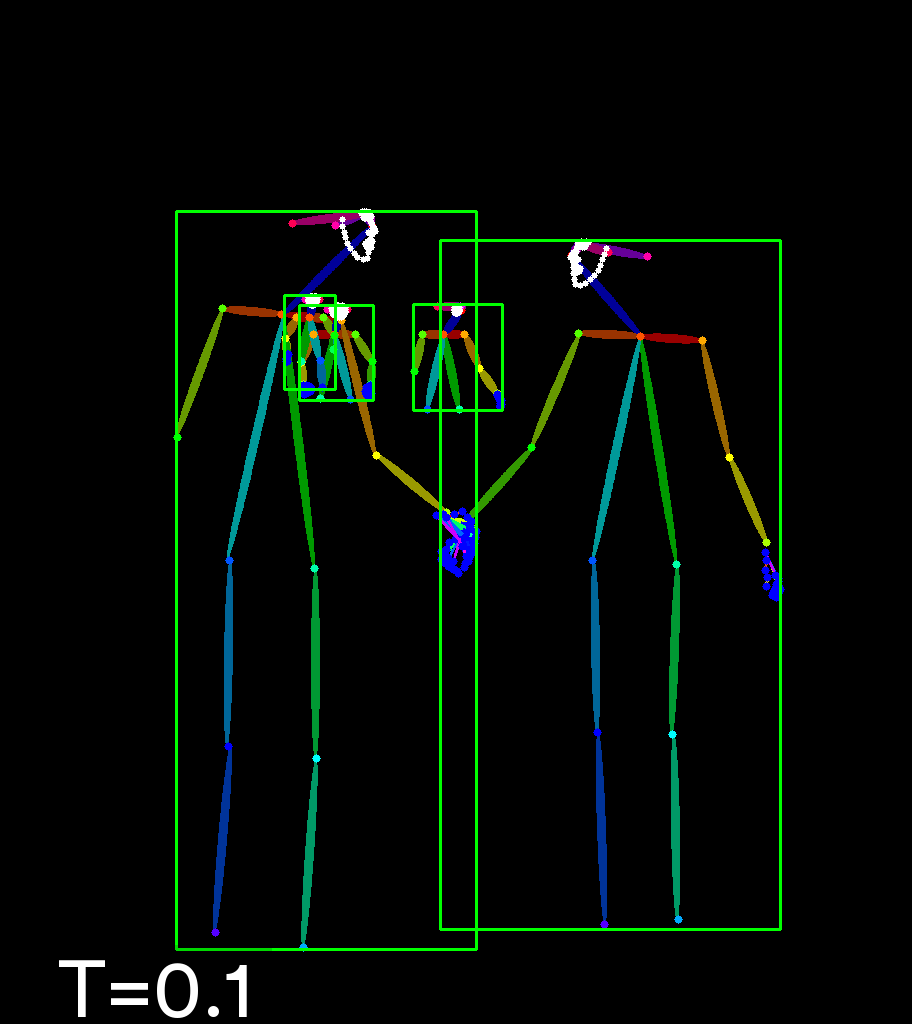}
\end{subfigure}
\begin{subfigure}{.32\textwidth}
  \centering
  \includegraphics[width=0.9\linewidth]{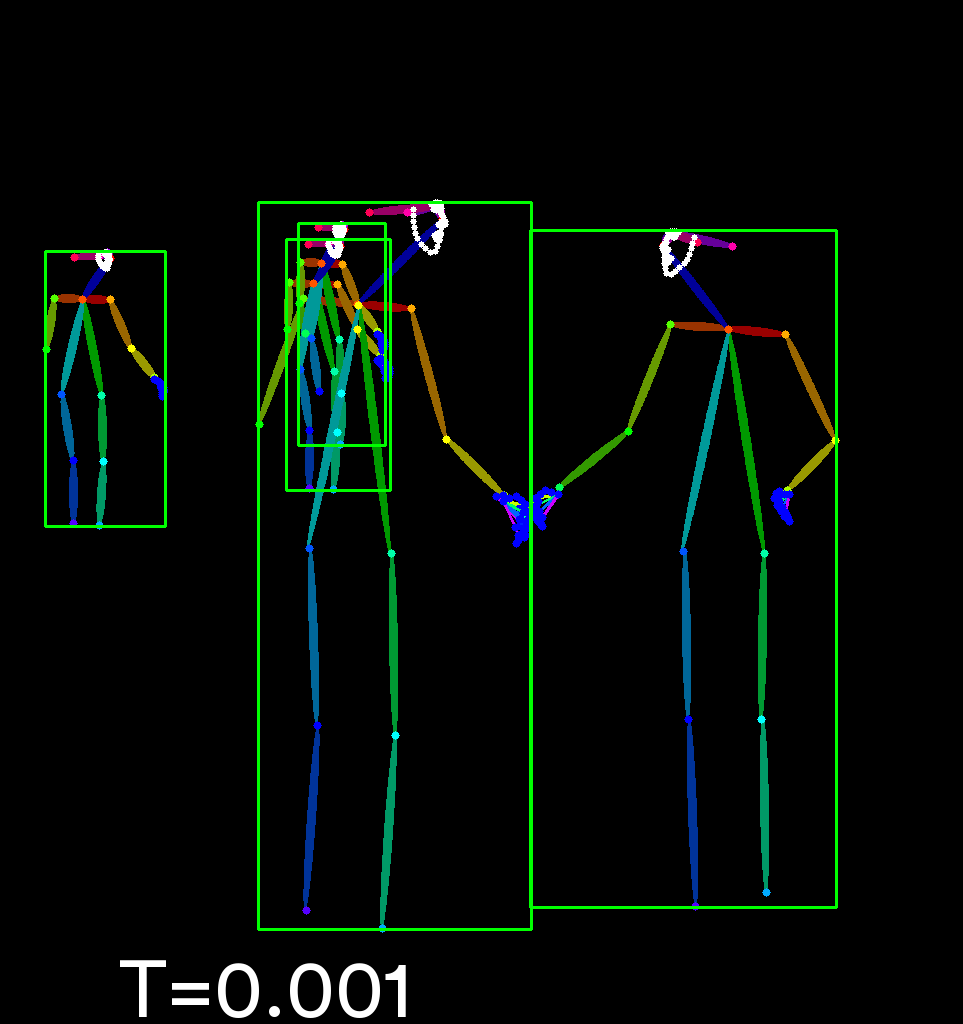}
\end{subfigure}
\caption{Effect of the temperature on GMM tempered sampling for text-to-pose generation. Prompt: ``two politicians shaking hands in a lobby''.}
    \label{fig: temperate sampling for text-to-pose}
\end{figure}

Tempered sampling for a GMM is crucial for our T2P model because the model with a temperature of 1 has too high entropy and ends up generating out-of-distribution high-noise pose sequences.
The effect of the temperature on the T2P inference can be observed in figure~\ref{fig: temperate sampling for text-to-pose} with the poses failing to shake hands at high temperature but eventually succeeding if the temperature is lowered.

\subsection{Pose adapter}
\begin{figure}[h!]
    \centering
    \includegraphics[width=\textwidth]{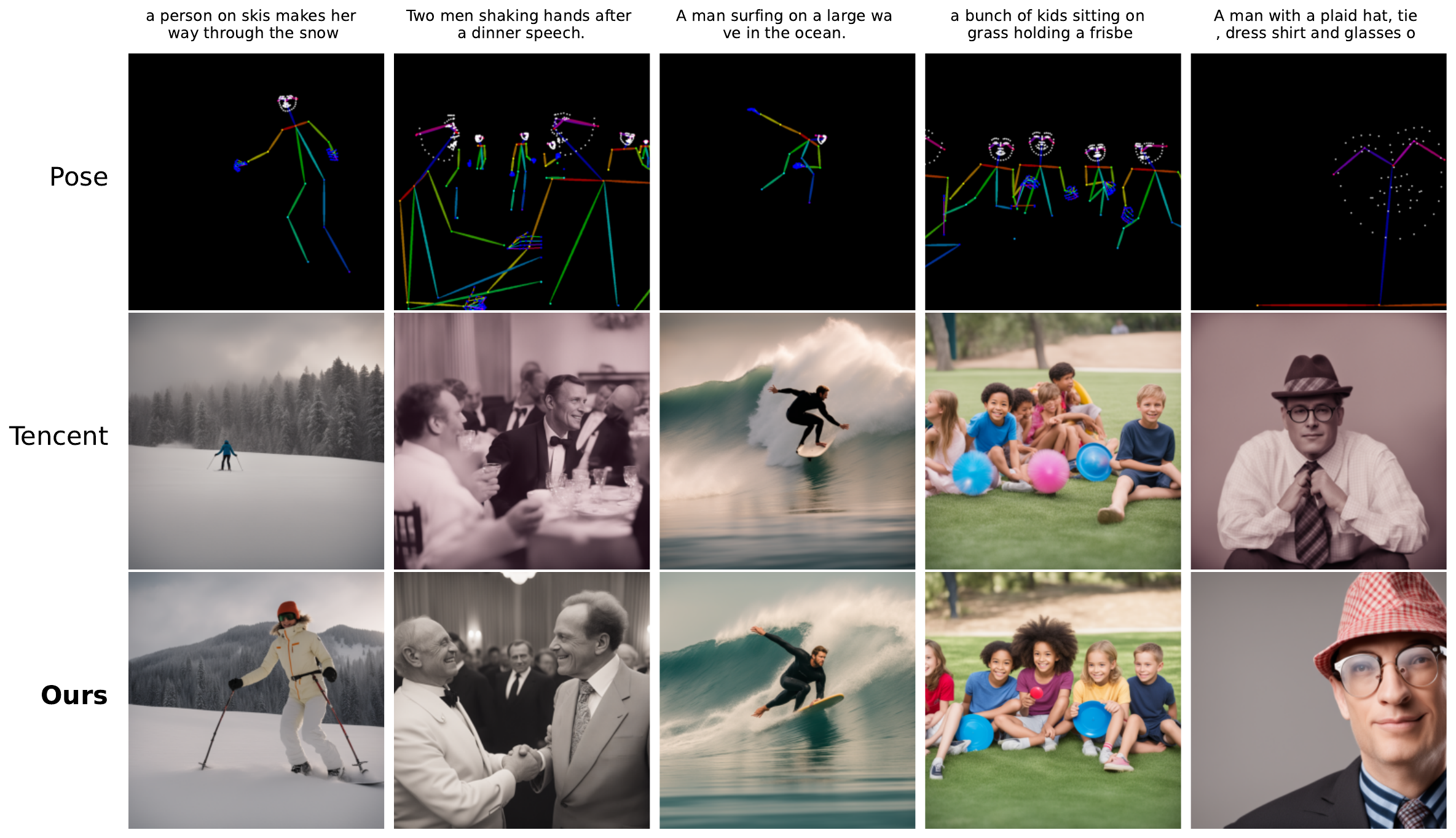}
    \caption{Pose-conditioned image generation using the SDXL-Tencent adapter, and ours (which includes hands and faces).}
    \label{fig: SDXL-Tencent adapter versus ours Dance}
\end{figure}

We train our adapter on high-quality images annotated with full poses (including bodies, hands, and faces).
This leads to a pose-conditioned image generation of higher quality and fidelity as shown in figure~\ref{fig: SDXL-Tencent adapter versus ours Dance}.

\subsection{Limits}

\paragraph{CLaPP metric}
The metric we designed to automatically assess the quality of generated poses has some shortcomings.
First, it was trained on a not-so-large dataset which means it may consider out-of-distribution captions or poses that are quite different from the dataset it was trained on, biasing a text-pose matching score.
Then, we used CLIP as a backbone, whose text and image encoders may have some pose-agnostic representations that hinder the quality of embeddings as much as human poses are concerned.
Finally, using images as representations of poses is highly inefficient, one could design a better model by working with the sequence of (x, y) points directly (much lower dimension), for instance using the backbone of the T2P model.

\paragraph{T2P}
Although the trained T2P model can generate rather precise poses, it tends to lack diversity and is really just a reflection of the data it was trained on.
Moreover, generating poses in an auto-regressive fashion is quite costly and adds considerable overhead at inference time.

\paragraph{Pose adapter}
Although our newly trained pose adapter has higher pose fidelity and image aesthetics, it does not always conform exactly to pose conditions and still suffers from lower image quality compared to the base SDXL.
More data would likely help mitigate both these issues.

\end{document}